\documentclass[letterpaper, 10 pt, conference]{ieeeconf}  

\IEEEoverridecommandlockouts                              

\usepackage{graphics} 
\usepackage{threeparttable}
\usepackage{times}
\usepackage{epsfig}
\usepackage{graphicx}
\usepackage{amsmath}

\usepackage{amssymb}
\usepackage{booktabs}
\usepackage[normalem]{ulem}
\usepackage{multirow}
\usepackage{colortbl}
\usepackage{color,soul}
\usepackage{bbding}
\usepackage{pifont}
\usepackage{wasysym}
\usepackage{arydshln}
\usepackage{subfigure}
\usepackage{hyperref}
\usepackage{graphicx}
\usepackage{times}
\usepackage{epsfig}
\usepackage{graphicx}
\usepackage{helvet}
\usepackage{courier}
\usepackage{float}
\usepackage{xcolor}

\definecolor{lightblue}{RGB}{173, 216, 230}
\usepackage{arydshln}
\usepackage{subfigure}
\usepackage[normalem]{ulem}
\usepackage{soul}
\soulregister\cite7 
\soulregister\citep7 
\soulregister\citet7 
\soulregister\ref7 
\soulregister\pageref7 
\usepackage{amsmath,amsfonts}
\usepackage{algorithmic}
\usepackage{algorithm}
\usepackage{array}
\usepackage{textcomp}
\usepackage{stfloats}
\usepackage{url}
\usepackage{verbatim}
\usepackage{graphicx}
\usepackage{arydshln}
\usepackage{bbding}
\usepackage{wrapfig}
\usepackage{caption}
\usepackage{newfloat}
\usepackage{listings}
\usepackage{xcolor}
\usepackage{graphicx}
\usepackage{multirow}
\usepackage{array}
\usepackage{hyperref} 
\newcommand{\redxmark}{\textcolor{red}{\ding{55}}}
\newcommand{\bluecheckmark}{\textcolor{blue}{\ding{51}}}

\title{\LARGE \bf
Complementary Information Guided Occupancy Prediction \\ via Multi-Level Representation Fusion
}

\author{Rongtao Xu, Jinzhou Lin, Jialei Zhou, Jiahua Dong, Changwei Wang, \\Ruisheng Wang, Li Guo, Shibiao Xu$^{\dag}$, Xiaodan Liang}

\begin{document}

\maketitle
\renewcommand{\thefootnote}{}
\footnotetext{
Rongtao Xu and Jinzhou Lin contributed equally.

$^{\dag}$Shibiao Xu is the corresponding author (shibiaoxu@bupt.edu.cn).

Rongtao Xu and Changwei Wang are with the Institute of Automation, Chinese Academy of Sciences, China. Jialei Zhou is with the Tongji University, China. Jiahua Dong is with the Shenyang Institute of Automation, Chinese Academy of Sciences, China. Jinzhou Lin, Li Guo, and Shibiao Xu are with the Beijing University of Posts and Telecommunications, China. Ruisheng Wang is with the University of Calgary, Canada. Xiaodan Liang is with Sun Yat-Sen University, China. 

This work was supported by the Beijing Natural Science Foundation (No.JQ23014), National Natural Science Foundation of China (No.62271074). 

}

\thispagestyle{empty}
\pagestyle{empty}

\begin{abstract}
Camera-based occupancy prediction is a mainstream approach for 3D perception in autonomous driving, aiming to infer complete 3D scene geometry and semantics from 2D images. Almost existing methods focus on improving performance through structural modifications, such as lightweight backbones and complex cascaded frameworks, with good yet limited performance. Few studies explore from the perspective of representation fusion, leaving the rich diversity of features in 2D images underutilized. Motivated by this, we propose \textbf{CIGOcc}, a two-stage occupancy prediction framework based on multi-level representation fusion. \textbf{CIGOcc} extracts segmentation, graphics, and depth features from an input image and introduces a deformable multi-level fusion mechanism to fuse these three multi-level features. Additionally, \textbf{CIGOcc} incorporates knowledge distilled from SAM to further enhance prediction accuracy. Without increasing training costs, CIGOcc achieves state-of-the-art performance on the SemanticKITTI benchmark. The code is provided in the supplementary material and will be released \href{https://github.com/VitaLemonTea1/CIGOcc}{\textcolor{magenta}{project page}}.
\end{abstract}
\section{Introduction}
Semantic Scene Completion (SSC), emerging as a promising solution for 3D perception, has recently played a crucial role in various applications within autonomous driving and robotics~\cite{xu2023scd,han2025multimodal,ma2025phyblock}. Camera-based 3D occupancy prediction is increasingly becoming a key and mainstream technology in SSC due to its high cost-effectiveness. However, this technology is currently struggling with accurately reconstructing occluded regions and maintaining cross-camera geometric consistency, limiting its ultimate performance from meeting expectations.

Although existing works~\cite{li2022bevformer,li2023voxformer,li2023fb} have achieved impressive performance, most primarily focus on optimizing network architectures, neglecting the adequate   exploration and utilization of image information at various levels. Consequently, these methods fail to deliver a more holistic and deeper recognization of 2D images, resulting in suboptimal 3D reconstruction. Specifically, these methods predominantly focus on graphics features such as position, size, color, and shape, which provide only partial semantics and represent mid-level features. However, the core of 3D perception lies in comprehending the spatial relationships in three dimensions. Depth maps, as carriers of distortion and depth information, naturally enhance the model's  ability to comprehend these relationships. Despite the fact that depth features carry little semantic information and are considered low-level features, their inclusion is crucial. Meanwhile, the rapid advancement of large foundational models has significantly boosted various downstream tasks. The pretrained SAM~\cite{sam_hq} with its strong semantic representations, can assist lightweight models more effectively capture image semantics and provide semantically-rich high-level segmentation features. Therefore, the skillful incorporation of foundational model representations and knowledge can be highly beneficial~\cite{xu2024mrftrans}. 

Therefore, the key challenge is how to effectively leverage low-level depth features and high-level segmentation features as complementary information to guide and enhance mid-level graphics features thereby improving the model’s recognization of 2D images.

\begin{figure}[!t]
    \centering
    \includegraphics[width=\columnwidth]{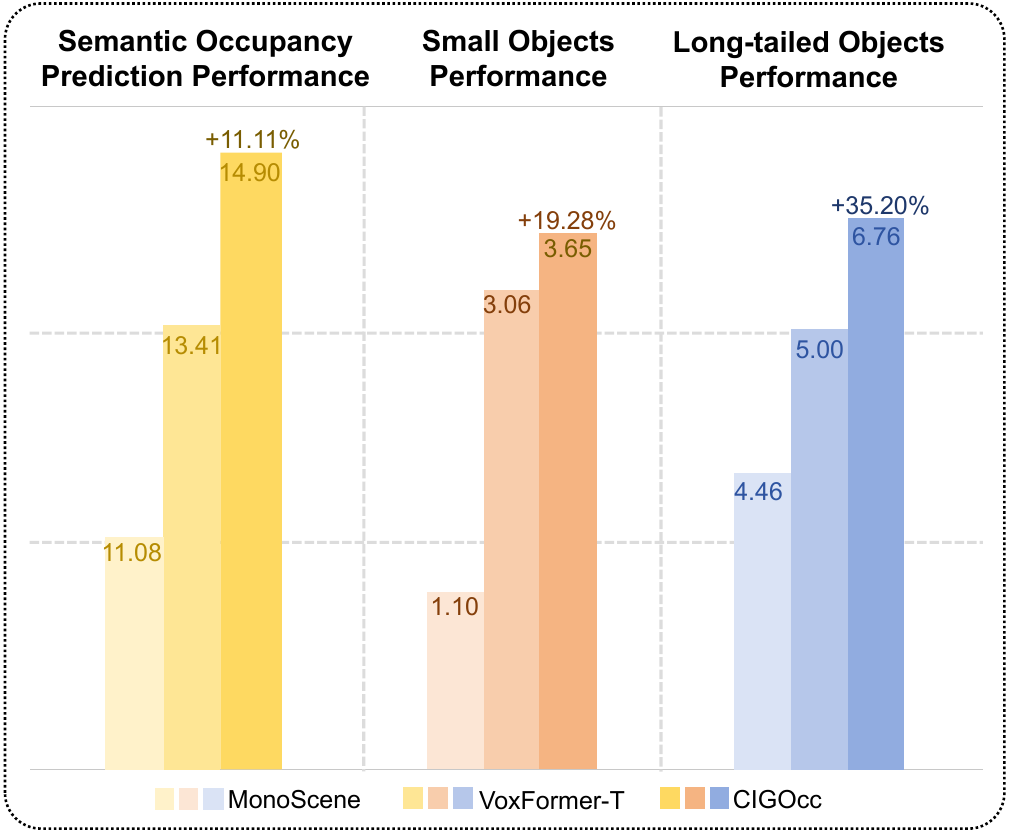} 
	 \caption{Quantitative results of semantic occupancy prediction performance, small objects performance and long-tailed performance. Compared to VoxFormer-T, our model has a significant improvement in mIoU.}
	 \label{fig:Display}
  \vspace{-0.9cm}
\end{figure}

To address this challenge, we propose a novel two-stage multi-level representation fusion network: \textbf{C}omplementary \textbf{I}nformation \textbf{G}uided \textbf{Occ}upancy (\textbf{CIGOcc}). In the first stage, we design a deformable multi-level fusion mechanism that conducts representation fusion of segmentation features and depth features from the input image. These two features, representing high-level qualitative information and low-level quantitative information respectively~\cite{xu2024deffusion}, exhibit the greatest disparity and provide the most complementary information to each other. In the second stage, we distill knowledge from Grounded-SAM~\cite{ren2024groundedsam} to enhance graphics features. The fused representation from the first stage is then used as complementary information to guide the second fusion and is fused with the graphics features. Finally, the resulting fused representation is used for occupancy prediction, outputting a voxel map.

Extensive experiments demonstrate the effectiveness of our method. Our contributions are threefold:

\begin{itemize}
\item CIGOcc Framework: We introduce a novel two-stage framework that utilizes multi-level representation fusion across diverse features to address the issue of low target precision and enable accurate 2D-to-3D reconstruction, particularly at greater distances.

\item Deformable Multi-Level Fusion Mechanism: We propose a new fusion mechanism that adaptively and effectively fuses depth and semantic information, ensuring a more comprehensive and accurate 3D reconstruction.

\item State-of-the-Art Performance: Our method achieves state-of-the-art performance in camera-based SSC, demonstrating its effectiveness and robustness in complex real-world scenarios.
\end{itemize}

\section{Related Work}

\subsection{Semantic Scene Completion}
SSC~\cite{SSC} is a crucial task in the field of autonomous driving and Embodied AI~\cite{xu2025a0,xu20253d,zhang2025robridge,zhou2025mathcal,zhang2025activevln}, aiming to enhance the vehicle's understanding of its surrounding environment by predicting the complete 3D structure of the scene and providing semantic labels for each voxel. Since SSC is not constrained by the inherent limitations of sensing resolution, occlusions, and incomplete observations from available sensors, it jointly infers complete scene geometry and
semantics from limited and often fragmented sensor data. As a result, SSC becomes the most promising solution for 3D perception~\cite{wang2021learning,popovic2021volumetric}, thus assisting vehicles in safe navigation and decision-making in complex and dynamic environments~\cite{han2024roomtour3d}.

Recently, various methods have been proposed to unlock the potential of SSC. For instance, SSCNet~\cite{SSC} utilizes 3D Convolutional Neural Networks (CNNs) to process sparse depth maps into dense 3D voxel grids and perform semantic labeling. EsscNet~\cite{essc} enhances SSC by integrating multi-scale features, allowing the network to capture both fine-grained and global contextual information. Some studies have applied Transformer architectures to SSC, using attention mechanisms to better capture long-range dependencies and complex contextual information within the scene. For instance, VoxFormer~\cite{li2023voxformer} employs a two-stage framework to elevate images to complete 3D voxelized semantic scenes.

\subsection{Camera-based 3D Perception}
Camera-based 3D perception is an important mode of 3D perception, aiming to extract three-dimensional information from two-dimensional images captured by cameras~\cite{xu2023rssformer}. Compared to other modes, such as LiDAR~\cite{lai2022stratified}, the camera-based mode can achieve good performance without high costs and has become a hot topic~\cite{lin2023development}.

Researchers have developed various methods to improve the accuracy and reliability of camera-based 3D perception. One fundamental method is monocular depth estimation. For example, Monodepth~\cite{monodepth} and Monodepth2~\cite{monodepth2} use CNNs to predict depth maps from single images. These models are trained on stereo image pairs, allowing them to learn the disparity between images and infer depth. Another noteworthy approach is the Detection Transformer (DETR) model~\cite{detr}. It uses attention mechanisms to enhance the accuracy of object detection in images. By incorporating the transformer architecture, DETR can simultaneously capture both local and global information within images, achieving better performance in complex visual tasks~\cite{li2022bevformer,tpvformer}~\cite{zhang2024navid}.

\subsection{3D Occupancy Prediction}
3D occupancy prediction is a core technology for realizing 3D perception. It reconstructs 3D scene structures from images by accurately predicting the occupancy of each voxel in 3D space using visual data~\cite{tang2024sparseocc}.  

Most of the existing studies predominantly utilize Transformer architectures. For example, VoxFormer~\cite{li2023voxformer} generates occupancy predictions through a two-stage architecture, resulting in producing detailed and accurate 3D occupancy maps. The other works have also boosted 3D occupancy prediction. For example, FB-Occ~\cite{li2023fb} combines Lift-Splat-Shoot (LSS) and BEVFormer~\cite{li2022bevformer} for bidirectional feature processing to effectively handle both bird's-eye view and front-view data, providing comprehensive scene understanding and improving prediction accuracy.

Although the above methods have achieved impressive performance in 3D occupancy prediction, they still do not fully exploit various features of images and do not consider further developing models' ability to recognise 2D images from the perspective of multi-level representation fusion~\cite{wang2024eaco}.
\begin{figure*}[ht]
    \centering
    \includegraphics[width=\textwidth]{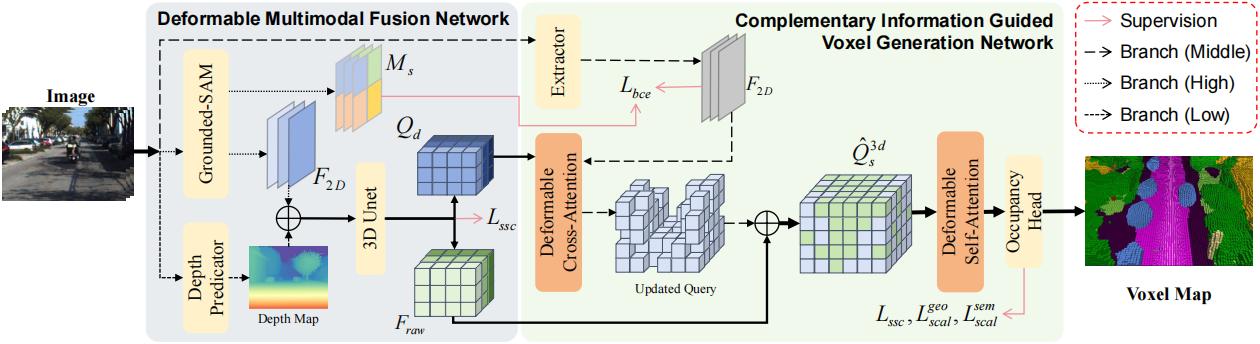} 
    \caption{Framework of CIGOcc. The input image is processed by Grounded-SAM to generate semantic features and segmentation masks, while the depth estimation network produces a depth map. DMFNet fuses the depth map and semantic features to generate initial voxel features and query proposals. For CIGNet, the image features extracted by ResNet, along with the query proposals, generate the voxel space via deformable cross-attention, which is then fused with DMFNet voxel features and enhanced through self-attention. Finally, the Occupancy Head performs occupancy prediction.}
    \label{fig:pipeline}
    \vspace{-0.3cm}
\end{figure*}

\section{Method}

The overall framework of CIGOcc is shown in the Fig.\ref{fig:pipeline}. CIGOcc consists of two stage: Deformable Multimodal Fusion Network (DMFNet) and Complementary Information Guided
Voxel Generation Network (CIGNet). DMFNet extracts high-level segmentation features and low-level depth features and performs representation fusion on them. CIGNet extracts mid-level graphics features, which will be enhanced by the complementary information and the knowledge distilled from Grounded-SAM. CIGNet also conducts representation fusion on complementary information and graphical information.


\subsection{Deformable Multi-Level Fusion Network}
Due to the powerful feature extraction capabilities of large vision models, and their rich prior knowledge, which excel in handling complex scenes and detail-rich images, we have incorporated Ground-SAM into the first part. Our first stage of training constructs the initial voxel space based on depth \(D_i \in \mathbb{R}^{C \times H \times W}\) from  and image semantic features \(F_{i}\in \mathbb{R}^{C \times H \times W }\) , while using Voxformer~\cite{li2023voxformer} to determine which voxels are worth focusing on and which can be separated as empty voxels.

Given 2D RGB image observations, we first generate stero depth estimates using the pre-trained binocular depth estimation network MobileStereoNet~\cite{shamsafar2022mobilestereonet}, which are then back-projected into point clouds. However, the voxel space generated from these point clouds \(P_c \in \mathbb{R}^{C \times H \times W}\) is of lower quality, especially at greater distances. Therefore, we embed semantic features extracted by Grounded-SAM to improve the quality of the voxel space constructed based on depth estimates.To fully leverage semantic features within the images, we additionally generated segmentation mask tokens \(M_s\) encoding object-specific information using Grounded-SAM during the second  stage of training.

To further enhance the quality of the voxel space, we propose DMFNet, a method adapted from LMSCNet~\cite{lmscnet}. Specifically, the initial point cloud information is fused with image features extracted by Grounded-SAM, followed by a lightweight Unet that transfers the 2D information into 3D space, enabling the extraction and fusion of multi-level features~\cite{xu2021dc}. This is then used to initially construct the voxel space through a 3D convolution layer:
\begin{equation}
    F_{raw}= \text{DMF}\left(F_i^{C \times H \times W}, D_{i}^{C \times H \times W}\right).
\end{equation}
Finally, an N-class segment head is applied to segment \(F_{raw}^{C \times H \times W \times D} \) into \(F_{seg}^{C_N \times H \times W \times D}\) , where each channel corresponds to a class occupancy prediction:
\begin{equation}
    F_{seg} = \text{SegHead}\left(F_{raw}\right).
\end{equation}
In the formula, \(C\), \(H\), \(W\) and \(D\) represent the channels, height, width, and depth, respectively, while \(C_N\) represents the N-class channels.To retain more rich and complete abstract feature information, we preserve \(F_{raw}\) for the second stage of training. \(F_{seg}\) is only used for the loss function calculation in the first stage.

Additionally, following VoxFormer, we obtained a total of \(N_d\) binary classification queries \(Q_d\) using LMSCNet, where each voxel is marked as 1 if it is occupied by at least one point. \(Q_d\) will be used as mask indices  during the second stage of training.

In the first stage, we mainly fused representations from different levels through DMFNet. By performing an initial occupancy prediction, we generated the coarse voxel space \(F_{raw}\). This approach can (i) \textbf{enhance feature representation with lower training costs by leveraging pre-trained large-scale vision models}, and (ii) \textbf{improve the quality of the coarse voxel space by correcting depth through image semantic features}~\cite{xu2022instance}.

\subsection{Complementary Information Guided Voxel Generation Network}
Previous occupancy prediction works have not used or referenced large vision models. To leverage the strong visual understanding capabilities of large vision models, we propose a method to distill Grounded-SAM into the occupancy prediction task. Additionally, to address the high computational complexity of traditional attention mechanisms when processing high-resolution images and long sequences, we adopt the deformable attention mechanism~\cite{zhu2020deformable} to construct the network.

Building on the first stage, we use the Resnet50 backbone~\cite{he2016deep} to extract image features \(F_{2D} \in \mathbb{R}^{\times H \times W \times D}\). Subsequently, to generate voxel features, we employed a two-step deformable attention mechanism similar to VoxFormer.

\begin{figure*}[!ht]
    \centering
    \includegraphics[width=\textwidth]{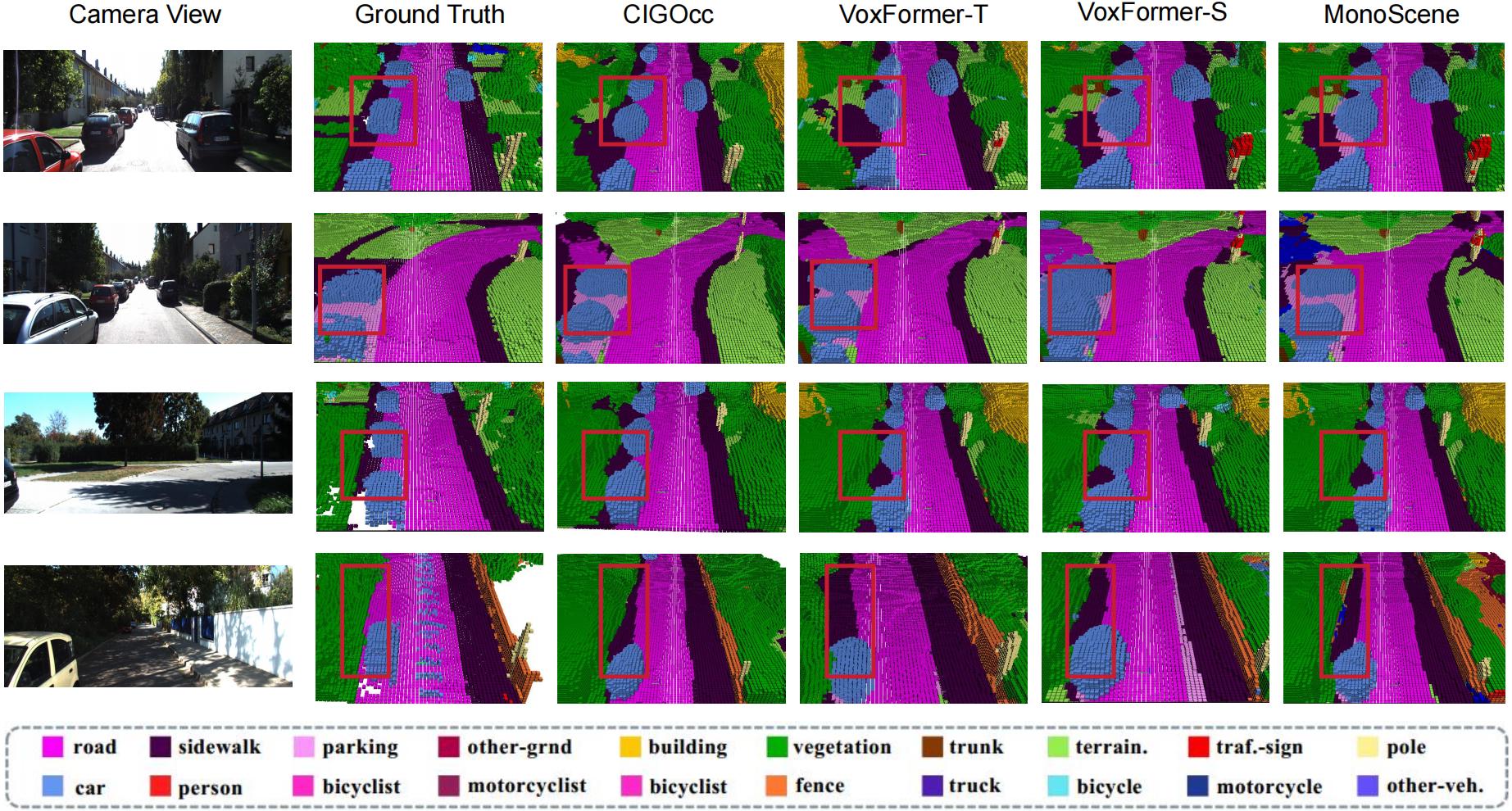} 
    \caption{Qualitative results of our method and others. We performed a visual comparison with three other models, and it can be seen that our model achieves more precise segmentation of scene voxels with less voxel overlapping, while also being more accurate in road prediction.}
    \label{fig:vis}
    \vspace{-0.3cm}
\end{figure*}

\textbf{Deformable cross-attention.} We utilized the binary classification queries \( Q_d\) obtained from previous stage as guiding indices. By leveraging the Deformable Cross-Attention mechanism (DCA), we embedded the 2D features \(F_{2D}\) into the 3D space \(Q_{s}^{3d}\), effectively guiding the representation transformation and construction of 3D space : 
\begin{equation}
    Q_{s}^{3d}= \textbf{DCA}(F_{2D},Q_d).    
\end{equation}

\textbf{Deformable self-attention.} To refine voxel features and enhance representational capacity, we initialize a voxel space and fuse the \(F_{raw}\) obtained from the first stage into it along with \(Q_{s}^{3d}\), thereby obtaining a multi-level voxel space. Simultaneously, we add mask tokens \(Mask \in \mathbb{R}^d\) based on \( Q_d\) to the voxel space to complete the scenes \(\hat{Q}_{s}^{3d}\). Then, by utilizing Deformable Self-Attention mechanism (DSA), we update the completed voxel space that will be used for prediction:

\begin{equation}
    \hat{V}_{s}^{3d}= \textbf{DSA}(\hat{Q}_{s}^{3d},\hat{Q}_{s}^{3d}).   
\end{equation}

Finally, we obtained the semantic voxel map \( Y_{t}^{C\times X\times Y\times Z} \) by up-sampling and linear mapping of the voxel space, where \( x \), \( y \), and \( z \) represent the 3D volume dimensions, and \( c \) represents the number of classes.

\textbf{Distillation module.} To distill the knowledge from Grounded-SAM into the model, we introduced a semantic decoder \( \theta_s \). The input to the semantic decoder is \(F_{2D}\) , with the segmentation mask tokens generated by Grounded-SAM in the previous stage serving as ground truth.
\begin{equation}
    F_{sem}^{2d}= \theta_s(F_{2D}).  
\end{equation}

We use binary cross-entropy loss to compute the difference between the predicted results and the mask tokens\(M_{s}\), in order to optimize the network.

In the second stage, we apply a lightweight deformable attention method and use the \(F_{raw}\)  to enhance our \(Q_{s}^{3d}\). \textbf{We distill the knowledge from the large-scale vision model to improve the model's semantic understanding, ensuring that the model performance is maximized without further increasing its size}.

\subsection{Training Loss}
In the first stage, we adopted a weighted cross-entropy loss from MonoScene\cite{cao2022monoscene}. It can be computed by:
\sloppy
\begin{equation}
    L_{ssc}= -\sum_{k=1}^{K}\sum_{c=c_0}^{c_M} w_{c}\hat{y}_{k,c}\log\left(\frac{e^{y_{k,c}}}{\sum_{c} e^{y_{k,c}}}\right).
\end{equation}
where \(k\) is the voxel index, \(K\) is the total number of the voxel, \(c\) indexes class, \(y_{k,c}\) is the predicted logits for the \(k\)-th voxel belonging to class \(c\), \(\hat{y}_{k,c}\) is the \(k\)-th element of ground truth voxel grid and is a one-hot vector (\(y_{i,k,c}\) = 1) if voxel \(k\) belongs to class \(c\)). \(w_{c}\) is a weight for each class according to the inverse of the class frequency as in~\cite{lmscnet}.

In the second stage, we used multiple loss functions:
\begin{enumerate}
\item For the distillation module, we used binary cross-entropy loss \(L_{bce}\) as distillation loss.
\item For the final output semantic voxel map, following MonoScene, we used the loss functions \(L_{scal}^{geo}\), \(L_{scal}^{sem}\), and \(L_{ssc}\)~\cite{cao2022monoscene}.
\end{enumerate}

The total loss function for the second stage is expressed as:
\begin{equation}
    L= \lambda_1L_{bce}+\lambda_2L_{scal}^{geo}+\lambda_3L_{scal}^{sem}+\lambda_4L_{ssc},    
\end{equation}
where \(\lambda_{1234}\) represent hyper-parameters.
\section{Experiment}

\subsection{Experimental Setup}
\sloppy
\textbf{Dataset.} We test the CIGOcc on the SemanticKITTI\cite{behley2019kitti} dataset, which provides dense semantic occupancy annotations for all LiDAR scans from the KITTI Odometry Benchmark. Each LiDAR scan covers a region extending from 0 to 51.2 meters in front of the vehicle, from -25.6 to 25.6 meters laterally, and from -2 to 4.4 meters in height. The ground truth is represented as a 256x256x32 3D voxel grid with a resolution of 0.2 meters per voxel. Each voxel is annotated as one of 20 classes. The dataset is divided into training, testing, and validation sets according to the official splits, and we report the results on the test set.

\begin{table*}[!ht]
\centering
\renewcommand{\arraystretch}{1.2}
\caption{Comparison with other camera-based methods.}
\label{table:comparison}
\resizebox{\textwidth}{!}{%
\begin{tabular}{|c|c|ccccccccccccccccccc|c|}
\hline
& &\multicolumn{18}{c}{\textbf{Semantic Occupancy Prediction}} &  \\ 
Method&Input&
\rotatebox{90}{\textcolor{magenta}{\rule{0.3cm}{0.3cm}} road (15.30\%)} & 
\rotatebox{90}{\textcolor{pink}{\rule{0.3cm}{0.3cm}} sidewalk (11.13\%)} & 
\rotatebox{90}{\textcolor{lightgray}{\rule{0.3cm}{0.3cm}} parking (1.12\%)} & 
\rotatebox{90}{\textcolor{orange}{\rule{0.3cm}{0.3cm}} other-ground (0.56\%)} & 
\rotatebox{90}{\textcolor{brown}{\rule{0.3cm}{0.3cm}} building (14.4\%)} & 
\rotatebox{90}{\textcolor{cyan}{\rule{0.3cm}{0.3cm}} car (3.92\%)} & 
\rotatebox{90}{\textcolor{teal}{\rule{0.3cm}{0.3cm}} truck (0.16\%)} & 
\rotatebox{90}{\textcolor{blue}{\rule{0.3cm}{0.3cm}} bicycle (0.03\%)} & 
\rotatebox{90}{\textcolor{purple}{\rule{0.3cm}{0.3cm}} motorcycle (0.03\%)} & 
\rotatebox{90}{\textcolor{darkgray}{\rule{0.3cm}{0.3cm}} other-vehicle (0.20\%)} & 
\rotatebox{90}{\textcolor{green}{\rule{0.3cm}{0.3cm}} vegetation (39.3\%)} & 
\rotatebox{90}{\textcolor{olive}{\rule{0.3cm}{0.3cm}} trunk (0.51\%)} & 
\rotatebox{90}{\textcolor{lime}{\rule{0.3cm}{0.3cm}} terrain (9.17\%)} & 
\rotatebox{90}{\textcolor{magenta}{\rule{0.3cm}{0.3cm}} person (0.07\%)} & 
\rotatebox{90}{\textcolor{red}{\rule{0.3cm}{0.3cm}} bicyclist (0.07\%)} & 
\rotatebox{90}{\textcolor{gray}{\rule{0.3cm}{0.3cm}} motorcyclist (0.05\%)} & 
\rotatebox{90}{\textcolor{lightblue}{\rule{0.3cm}{0.3cm}} fence (3.90\%)} & 
\rotatebox{90}{\textcolor{yellow}{\rule{0.3cm}{0.3cm}} pole (0.29\%)} & 
\rotatebox{90}{\textcolor{orange}{\rule{0.3cm}{0.3cm}} traffic-sign (0.08\%)} &
 mIoU
\\ \hline
LMSCNet\cite{lmscnet} & Camera & 46.70 & 19.50 & 13.50 & 3.10 & 10.30 & 14.30 & 0.30 & 0.00 & 0.00 & 0.00 & 10.80 & 0.00 & 10.40 & 0.00 & 0.00 & 0.00 & 5.40 & 0.00 & 0.00 & 7.07\\ 
3DSketch\cite{chen20203dsketch} & Camera & 37.70 & 19.80 & 0.00 & 0.00 & 12.10 & 17.10 & 0.00 & 0.00 & 0.00 & 0.00 & 12.10 & 0.00 & 16.10 & 0.00 & 0.00 & 0.00 & 3.40 & 0.00 & 0.00&6.23 \\ 
AICNet\cite{li2020AIC} & Camera & 39.30 & 18.30 & 19.80 & 1.60 & 9.60 & 15.30 & 0.70 & 0.00 & 0.00 & 0.00 & 9.60 & 1.90 & 13.50 & 0.00 & 0.00 & 0.00 & 5.00 & 0.10 & 0.00 & 7.09 \\ 
JS3C-Net\cite{lmscnet}  & Camera & 47.30 & 21.70 & 19.90 & 2.80 & 12.70 & 20.10 & 0.80 & 0.00 & 0.00 & 4.10 & 14.20 & 3.10 & 12.40 & 0.00 & 0.20 & 0.20 & 8.70 & 1.90 & 0.30 & 8.97 \\ 
MonoScene\cite{cao2022monoscene}  & Camera & 54.70 & 27.10 & 24.80 & 5.70 & 14.40 & 18.80 & 3.30 & 0.50 & 0.70 & 4.40 & 14.90 & 2.40 & 19.50 & 1.00 & 1.40 & \underline{0.40} & 11.10 & 3.30 & 2.10 & 11.08 \\ 
OccFormer\cite{zhang2023occformer}  & Camera & 55.90 & \underline{30.30} & \textbf{31.50} & 6.50 & 15.70 & 21.60 & 1.20 & 1.50 & \underline{1.70} & 3.20 & 16.80 & 3.90 & 21.30 & 2.20& 1.10 & 0.20 & 11.90& 3.80 & 3.70 & 12.32 \\ 
SurroundOcc\cite{wei2023surroundocc}  & Camera & 56.90& 28.30 & \underline{30.20} & 6.80 & 15.20 & 20.60 & 1.40 & 1.60 & 1.20 & 4.40 & 14.90 & 3.40 & 19.30 & 1.40 & 2.00 & 0.10 & 11.30 & 3.90 & 2.40 & 11.86\\
TPVFormer\cite{tpvformer}  & Camera & 55.10 & 27.20 & 27.40 & 6.50 & 14.80 & 19.20 & 3.70 & 1.00 & 0.50 & 2.30 & 13.90 & 2.60 & 20.40 & 1.10 & \underline{2.40}& \underline{0.30} & 11.00 & 2.90 & 1.50 & 11.26 \\ 
SparseOcc\cite{tang2024sparseocc} &Camera&\textbf{59.59}&29.68 &20.44&0.47&15.41&24.03&\textbf{18.07}& 0.78& 0.89& \textbf{8.94}& 18.89& 3.46& 31.06& \textbf{3.68}& 0.62& 0.00& 6.73& 3.89& 2.60&13.12\\
MonoOcc-S\cite{zheng2024monoocc}&Camera &55.20 &27.80 &25.10 &\textbf{9.70} &21.40 &23.20 &5.20 &\underline{2.20} &1.50 &5.40 &24.00 &8.70 &23.00 &1.70 &2.00 &0.20 &\underline{13.40} &5.80 &6.40 &13.80\\
LowRankOcc\cite{Zhao_2024lowrankocc}&Camera&52.80 &27.20 &25.10& \underline{8.80}& 22.10& 20.90& 2.90& \textbf{3.30}& \textbf{2.70}& 4.40& 22.90& \underline{8.90}& 20.80& 2.40& 1.70& \textbf{2.30}& \textbf{14.40}& \underline{7.00}& \underline{7.00}&13.56\\
VoxFormer-S\cite{li2023voxformer}  & Camera & 53.90 & 25.30 & 21.10 & 5.60 & 19.80 & 20.80 & 3.50 & 1.00 & 0.70 & 3.70 & 22.40 & 7.50 & 21.30 & 1.40 & \textbf{2.60} & 0.00 & 11.10 & 5.10 & 4.90 & 12.20 \\ 
VoxFormer-T\cite{li2023voxformer}  & Camera & 54.10 & 26.90 & 25.10 & 7.30 & \underline{23.50} & 21.70 & 3.60 & 1.90 & 1.60 & 4.10 & 24.40 & 8.10& 24.20 & 1.60 & 1.10 & 0.00 & 13.10 & 6.60& 5.70 & \underline{13.41} \\ 
   \hline
DMFNet& Camera &55.25&25.02&3.06&0.00&17.90&\underline{26.76}&0.00&0.00&0.00&0.00&\underline{25.92}&0.05&\underline{28.44}&0.00&0.00&0.00&4.10&0.17&0.00&9.77 \\
CIGOcc & Camera  &\underline{57.12} & \textbf{30.53}&19.70&0.82&\textbf{24.12}&\textbf{28.56}   & \underline{11.84} &1.61 &  1.49 &\underline{7.63} & \textbf{26.96} & \textbf{8.95}&  \textbf{34.28} & \underline{2.53} & 1.05 & 0.00& 8.40 &\textbf{9.70} & \textbf{7.86} &\textbf{14.90}\\ \hline
\end{tabular}%
}

\begin{flushleft}
\small The best results are highlighted in \textbf{bold}, while the second-best results are \underline{underlined} for clarity.
\end{flushleft}
\vspace{-0.3cm}
\end{table*}

\textbf{Evaluation Metrics.} Similar to other works, we use mean Intersection over Union (mIoU) as the evaluation metric for semantic occupancy.

\subsection{Comparison with Other Methods and Results}

In the first stage of training, we chose the pre-trained weights ViT-H HQ-SAM~\cite{sam_hq} for Grounded-SAM and MSNet3D SFDS~\cite{shamsafar2022mobilestereonet} for MobileStereoNet, training for 20 epochs on 4 RTX 3090 GPUs, taking 4.5 hours. In the second stage, we used the ResNet50~\cite{resnet} backbone, training for 20 epochs on 4 RTX 3090 GPUs, which also took 4.5 hours. The specific comparison results are shown in Table \ref{table:comparison}.

We compared our method with other approaches using the SemanticKITTI dataset. Table \ref{table:comparison} includes semantic occupancy prediction methods based on camera and RGB images within a 51.2m range. To be specific, our method shows significant improvements in certain categories, and the mIoU surpasses all other baselines, setting a new state-of-the-art (SOTA). Table \ref{table:evalrange} presents a performance comparison of the model under different volumes (12.8x12.8x6.4\(m^3\), 25.6x25.6x6.4\(m^3\), 51.2x51.2x6.4\(m^3\)). It can be observed that not only in the 51.2m range, but also within the 12.8m and 25.6m ranges, the mIoU and IoU are higher than those of other models. Our model demonstrates a greater advantage in close-range scenarios compared to other models, which is more desirable in autonomous driving. This is because the model's accurate perception of close-range distances can improve its judgment of longer distances.


To ensure fairness, we conducted a detailed comparison between our method and VoxFormer-T. Since MonoOcc-L~\cite{zheng2024monoocc} uses its own pre-trained large backbone InterImage-XL~\cite{wang2022internimage}, we only compared with MonoOcc-S, which uses ResNet50. Overall, our method achieved a 1.49 \% improvement in mIoU, and it also showed significant improvements in most categories. For instance, long-tailed objects like \textit{truck} ($0.32\%$, 3.60~$\rightarrow$~11.84) and \textit{other-vehicle} ($0.2\%$, 4.10~$\rightarrow$~7.63), along with small objects such as \textit{person} ($0.07\%$, 1.60~$\rightarrow$~2.53) and \textit{traffic-sign} ($0.08\%$, 5.70~$\rightarrow$~7.86).

Table \ref{table:comparison} also presents the training results of DMFNet. The comparison of the two-stage results demonstrates that our second-stage is indeed effective. In particular, it achieved significant breakthroughs in some small objects and long-tailed objects, such as \textit{truck} and \textit{bicycle}.

As shown in the Fig. \ref{fig:vis}, we conducted a qualitative comparison between our method and other models. Our method demonstrates clearer segmentation, with less overlap between voxels of different classes.

    
    

\subsection{Ablation Study}
We conducted ablation experiments on the components of our method using the SemanticKITTI dataset. Each table provides detailed data on the independent impact of each component. It is worth noting that Grounded-SAM is only used to generate segmentation mask tokens and extract image features during the first stage of training.
\vspace{-1mm}
\begin{table}[h!]
    \centering
    \caption{Ablation Study of Semantic Auxiliary Loss}
    \label{table:ablation1}
    \resizebox{0.3\textwidth}{!}{%
    \begin{tabular}{c |cc}
    \hline
    Semantic auxiliary loss& mIoU \\
    \hline
    
    \redxmark  & 14.10 \\
    
    \bluecheckmark  &  14.49\\
    \hline

    \end{tabular}%
    }
\end{table}

\begin{table*}[ht!]
\centering
    \caption{Quantitative comparison on different volumes.}
    \label{table:evalrange}
    \renewcommand{\arraystretch}{1.2}
    \resizebox{\textwidth}{!}{%
\begin{tabular}{l|ccc|ccc|ccc|ccc}
\hline
\multicolumn{1}{c|}{Method}                                                                       & \multicolumn{3}{c|}{CIGOcc}                       & \multicolumn{3}{c|}{VoxFormer-T}                & \multicolumn{3}{c|}{VoxFormer-S}              & \multicolumn{3}{c}{MonoScene}                   \\ \hline
\multicolumn{1}{c|}{range}                                                                        & 12.8m          & 25.6m          & 51.2m          & 12.8m          & 25.6m          & 51.2m         & 12.8m         & 25.6m         & 51.2m         & 12.8m          & 25.6m          & 51.2m         \\ \hline
\multicolumn{1}{c|}{IoU(\%)}                                                                      & \textbf{67.66} & \textbf{59.04} & \textbf{44.28} & 65.38          & 57.69          & 44.15         & 65.35         & 57.54         & 44.02         & 38.42          & 38.55          & 36.80         \\
\multicolumn{1}{c|}{Precision(\%)}                                                                & \textbf{81.55} & \textbf{74.03} & \textbf{64.64} & 76.54          & 69.95          & 62.06         & 77.65         & 70.85         & 62.32         & 51.22          & 51.96          & 52.19         \\
\multicolumn{1}{c|}{Recall(\&)}                                                                   & 79.90          & 74.46          & 58.45          & 81.77          & 76.70          & 60.47         & { \underline{80.49} }   & { \underline {75.39} }  & { \underline {59.99} }  & 60.60          & 59.91          & 55.50         \\ \hline
\multicolumn{1}{c|}{mIoU}                                                                         & \textbf{23.81} & \textbf{20.35} & \textbf{14.90} & { \underline {21.55}}    & { \underline {18.42}}    & { \underline {13.35}}   & 17.66         & 16.48         & 12.35         & 12.25          & 12.22          & 11.30         \\
{\textcolor{cyan}{\rule{0.2cm}{0.2cm}} car 3.92\% }           & \textbf{48.00} & \textbf{39.47} & \textbf{28.56} & { \underline{ 44.90} }   & { \underline {37.46} }   & { \underline {26.54}}   & 39.78         & 35.24         & 25.79         & 24.34          & 24.64          & 23.29         \\
{\textcolor{blue}{\rule{0.2cm}{0.2cm}} bicycle 0.03\%}        & \textbf{5.43}  & \textbf{5.63}  & \textbf{1.61}  & { \underline {5.22}}     & { \underline {2.87} }    & { \underline {1.28}}    & 3.04          & 1.48          & 0.59          & 0.07           & 0.23           & 0.28          \\
{\textcolor{purple}{\rule{0.2cm}{0.2cm}} motorcycle 0.03\%}   & \textbf{7.82}  & \textbf{3.69}  & \textbf{1.49}  & { \underline {2.98}}     & { \underline {1.24}}     & 0.56          & 2.84          & 1.10          & 0.51          & 0.05           & 0.20           & { \underline {0.59}}    \\
{\textcolor{teal}{\rule{0.2cm}{0.2cm}} truck 0.16\% }         & { \underline {12.52}}    & { \underline {11.}}    & \textbf{11.84} & 9.80           & 10.38          & 7.26          & 7.50          & 7.47          & 5.63          & \textbf{15.44} & \textbf{13.84} & 9.29          \\
{\textcolor{darkgray}{\rule{0.2cm}{0.2cm}} other-veh.0.20\%}  & { \underline {11.77} }   & { \underline {5.81}}     & { \underline {7.63} }    & \textbf{17.21} & \textbf{10.61} & \textbf{7.81} & 8.71          & 4.98          & 3.77          & 1.18           & 2.13           & 2.63          \\
{\textcolor{magenta}{\rule{0.2cm}{0.2cm}} person 0.07\%}      & 3.31           & 2.76           & { \textbf {2.53} }    & \textbf{4.44}  & \textbf{3.50}  & \underline{1.93} & { \underline {4.10}}    & { \underline {3.31}}    & 1.78          & 0.90           & 1.37           & 2.00          \\
{\textcolor{red}{\rule{0.2cm}{0.2cm}} bicyclist 0.07\%}       & 0.86           & 2.43           & 1.05           & { \underline {2.65} }    & { \underline {3.92}}     & { \underline {1.97}}    & \textbf{6.82} & \textbf{7.10} & \textbf{3.32} & 0.54           & 1.00           & 1.07          \\
{\textcolor{gray}{\rule{0.2cm}{0.2cm}} motorcyclist 0.05\%}   & 0.00           & 0.00           & 0.00           & 0.00           & 0.00           & 0.00          & 0.00          & 0.00          & 0.00          & 0.00           & 0.00           & 0.00          \\
{\textcolor{magenta}{\rule{0.2cm}{0.2cm}} road 15.30\% }       & \textbf{79.99} & \textbf{71.52} & \textbf{57.12} & { \underline {75.45}}    & { \underline {66.15} }   & 53.57         & 72.40         & 65.74         & 54.76         & 57.37          & 57.11          & { \underline {55.89}}   \\
{\textcolor{lightgray}{\rule{0.2cm}{0.2cm}} parking 1.12\%}   & \textbf{30.82} & \textbf{28.79} & \textbf{19.70} & { \underline {21.01} }   & { \underline {23.96}}    & { \underline {19.69}}   & 10.79         & 18.49         & 15.50         & 20.04          & 18.60          & 14.75         \\
{\textcolor{pink}{\rule{0.2cm}{0.2cm}} sidewalk 11.13\%}      & \textbf{54.08} & \textbf{39.55} & \textbf{30.53} & { \underline {45.39}}    & { \underline {34.53}}    & { \underline {26.52}}   & 39.35         & 33.20         & 26.35         & 27.81          & 27.58          & 26.50         \\
{\textcolor{orange}{\rule{0.2cm}{0.2cm}} other-ground 0.56\%} & { \underline {0.13}}     & 0.13           & { \underline{ 0.82}  }   & 0.00           & 0.76           & 0.42          & 0.00          & { \underline {1.54}  }  & 0.70          & \textbf{1.73}  & \textbf{2.00}  & \textbf{1.63} \\
{\textcolor{brown}{\rule{0.2cm}{0.2cm}} building 14.4\%}      & \textbf{25.33} & \textbf{31.96} & \textbf{24.12} & { \underline {25.13} }   & { \underline{ 29.45}  }  & { \underline {19.54}}   & 17.91         & 24.09         & 17.65         & 16.64          & 15.97          & 13.55         \\
{\textcolor{lightblue}{\rule{0.2cm}{0.2cm}} fence 3.90\%}     & \textbf{19.80} & \textbf{14.00} & \textbf{8.40}  & { \underline {16.17}}    & { \underline {11.15} }   & 7.31          & 12.98         & 10.63         & { \underline{ 7.64} }   & 7.57           & 7.37           & 6.60          \\
{\textcolor{green}{\rule{0.2cm}{0.2cm}} vegetation 39.3\%}    & \textbf{46.81} & \textbf{40.20} & \textbf{26.96} & { \underline {43.55} }   & { \underline {38.07}}    & 16.10         & 40.50         & 34.68         & { \underline {24.39} }  & 19.59          & 19.68          & 17.98         \\
{\textcolor{olive}{ \rule{0.2cm}{0.2cm}} trunk 0.51\%}         & \textbf{24.47} & \textbf{16.11} & \textbf{8.95}  & { \underline {21.39}}    & { \underline {12.75}}    & { \underline {6.10}}    & 15.81         & 10.64         & 5.08          & 2.02           & 2.57           & 2.44          \\
{\textcolor{lime}{\rule{0.2cm}{0.2cm}} terrain 9.17\% }       & \textbf{49.67} & \textbf{44.99} & \textbf{34.28} & { \underline {42.82}}    & { \underline {39.64}}    & { \underline {33.06} }  & 32.25         & 35.08         & 29.96         & 31.72          & 31.59          & 29.84         \\
{\textcolor{yellow}{\rule{0.2cm}{0.2cm}} pole 0.29\%}         & \textbf{20.97} & \textbf{17.37} & \textbf{9.70}  & { \underline {20.66} }   & { \underline {15.56}}    & { \underline {9.15}}    & 14.47         & 11.95         & 7.11          & 3.10           & 3.79           & 3.91          \\
{\textcolor{orange}{\rule{0.2cm}{0.2cm}} traffic-sign 0.08\%} & \textbf{10.64} & \textbf{9.08}  & \textbf{7.86}  & { \underline {10.63}}    & { \underline {8.09}}     & { \underline {4.94} }   & 6.19          & 6.29          & 4.18          & 3.69           & 2.54           & 2.43          \\ \hline
\end{tabular}%
}
\vspace{-3mm}
\begin{flushleft}
\small For each range, the best results are highlighted in \textbf{bold}, while the second-best results are \underline{underlined} for clarity.
\end{flushleft}
\vspace{-0.3cm}
\end{table*}

\textbf{Semantic auxiliary loss:} We first performed an ablation study on the semantic decoder, particularly examining whether the Semantic Auxiliary Loss was used to distill Grounded-SAM knowledge into the second stage. Table \ref{table:ablation1} shows the detailed results. The results indicate that, compared to the complete model, there is a certain degree of decrease in mIoU. This demonstrates the feasibility and effectiveness of distilling knowledge from large vision models into the occupancy task in this manner.

\textbf{Fusion Feature:} Subsequently, we conducted an ablation study on the Fusion Feature (using Semantic Auxiliary Loss), where only depth was used to generate \(F_{raw}\) without incorporating features extracted by Gounded-SAM. The detailed results are shown in Table \ref{table:ablation2}.The results indicate that integrating features ensures a more comprehensive and accurate 3D scene reconstruction and it has a significant impact on the model.

\vspace{-1mm}
\begin{table}[h!]
    \centering
    \caption{Ablation Study of Fusion Feature}
    \label{table:ablation2}
    \resizebox{0.2\textwidth}{!}{%
    \begin{tabular}{ c |cc}
    \hline
     Fusion Feature& mIoU \\
    \hline
    \redxmark & 13.85 \\
     \bluecheckmark & 14.49 \\
    \hline
    
    \end{tabular}%
    }
    \vspace{-3mm}
\end{table}

\textbf{Grounded-SAM:} We conducted an ablation study on the entire Grounded-SAM model, where only depth was used to generate \(F_{raw}\) and without using the Semantic Auxiliary Loss. The detailed results are shown in Table \ref{table:ablation3}. Overall, the mIoU decreased by 0.86. Comparing this with other results, it can be observed that introducing large vision model into the occupancy task can effectively enhance the model's semantic understanding and scene reconstruction capabilities.

\vspace{-1mm}
\begin{table}[!]
    \centering
    \caption{Ablation Study of Grounded-SAM}
    \label{table:ablation3}
    \resizebox{0.25\textwidth}{!}{%
    \begin{tabular}{ c |cc}
    \hline
     Segment-Anything& mIoU \\
    \hline
    \redxmark & 13.63 \\
     \bluecheckmark & 14.49 \\
    \hline
    
    \end{tabular}%
    }
    
\end{table}

Based on the above, by incorporating Grounded-SAM and the DMFNet, we effectively improved the accuracy of the original method.
\subsection{Model efficiency}

We conducted a training consumption test on a single RTX 3090 GPU with a batch size of 1. Compared to VoxFormer-T, our training memory increased by 0.4G, latency increased by 0.03 seconds, and the total training time increased by one hour. Although there is a slight increase in training consumption, the improvement in mIoU is significantly greater than the increase in training consumption.
\vspace{-1mm}
\begin{table}[h!]
    \centering
    \caption{Model efficiency}
    \label{table:effi}
    \resizebox{0.5\textwidth}{!}{%
    \begin{tabular}{c| ccc}
    \hline
    Method& Latency(s) & Train MEM(G)& Total hours(h)\\
    \hline
    VoxFormer-T& 0.76 & 16.6G &16\\
    Ours & 0.79 &  17G&17\\

    \hline
    
    \end{tabular}%
}
\end{table}
\vspace{-5mm}
\section{Conclusion}
The proposed CIGOcc is a high-performance and efficient occupancy prediction framework. We introduce large vision model into the semantic occupancy task and improve existing semantic occupancy prediction method through semantic auxiliary loss and CIGNet. By incorporating large vision models, more comprehensive knowledge is transferred to the semantic occupancy task, enhancing the framework's performance while maintaining a balance in efficiency. Utilizing the method described in this paper, CIGOcc achieved SOTA performance on the SemanticKITTI dataset.

\bibliographystyle{IEEEtran}
\bibliography{References}

\end{document}